\documentclass{article}

\usepackage[nonatbib, final]{nips_2017}

\usepackage[utf8]{inputenc} 
\usepackage[T1]{fontenc}    
\usepackage{hyperref}       
\usepackage{url}            
\usepackage{booktabs}       
\usepackage{amsfonts}       
\usepackage{nicefrac}       
\usepackage{microtype}      
\usepackage{verbatim}

\usepackage{hyperref}
\usepackage{url}
\usepackage{amsmath,graphicx,booktabs,amsfonts}

\title{Improving   Classification  Rate   of  Schizophrenia   Using  a
Multimodal Multi-Layer Perceptron Model with Structural and Functional
MRI}

\author{   Alvaro  E.   Ulloa$^{1}$,   Sergey  Plis$^{2}$,   Vince
D.     Calhoun$^{1,2}$\\     \texttt{alvarouc@unm.edu, [splis,vcalhoun]@mrn.org}\\
$^1$Department of  Electrical and Computer  Engineering,The University
of New Mexico, NM USA\\  $^2$The Mind Research Network,
NM USA }

\begin{document}

\maketitle

\begin{abstract}

  The wide variety of brain imaging technologies allows us to exploit
  information inherent to different data modalities.  The richness of
  multimodal datasets may increase predictive power and reveal latent
  variables that otherwise would have not been found.  However, the
  analysis of multimodal data is often conducted by assuming linear
  interactions which impact the accuracy of the results.

  We propose the use of a multimodal multi-layer perceptron model to
  enhance the predictive power of structural and functional magnetic
  resonance imaging (sMRI and fMRI) combined.  We also use a synthetic
  data generator to pre-train each modality input layers, alleviating
  the effects of the small sample size that is often the case for
  brain imaging modalities.

  The proposed model improved the average and uncertainty of the area
  under the ROC curve to 0.850$\pm$0.051 compared to the best results
  on individual modalities (0.741$\pm$0.075 for sMRI, and
  0.833$\pm$0.050 for fMRI).

\end{abstract}

\section{Introduction}

Recent advances  on detecting and monitoring  biomedical signals makes
simultaneous investigation of multiple aspects of human physiology and
their interactions more practical  and efficient. This is particularly
true for studies  on brain imaging, where  functional, structural, and
chemical composition of the brain can be measured.

Brain  imaging is  often used  for evaluation  and research  of mental
disorders, mainly due  to its ability to  directly measure information
about the  brain.  Schizophrenia  is among  the most  prevalent mental
disorders,   affecting  about   1\%   of   the  population   worldwide
\cite{bhugra2005global}.  Schizophrenia is  a devastating disease that
alters normal  behavior and may provoke  hallucinations.  Depending on
the severity, schizophrenia can impair an individual and significantly
degrade his/her  quality of life,  resulting in a social  burden.  The
need to better diagnose and treat this disorder motivates the study of
schizophrenia at the behavioral and biological levels.

In  this paper,  we focus  on assessing  the predictive  power of  two
popular  brain  imaging   modalities:  structural  magnetic  resonance
imaging  (sMRI) and  functional MRI  (fMRI).  SMRI  is a  non-invasive
technique for measuring gray matter concentration (GMC) which provides
direct  anatomical  information  of  brain  structure  in  form  of  a
three-dimensional   image.     On   the   other   hand,    fMRI   uses
blood-oxygen-level  dependent contrast  to estimate  blood flow  as an
indirect measure  of neuronal  activity in the  brain over  time which
results in a four-dimensional image.

The literature presents overwhelming  evidence that GMC abnormalities,
captured  by  sMRI,  among individuals  diagnosed  with  schizophrenia
yields        meaningful        information        for        clinical
evaluation~\cite{meda2008large,                     gupta2014patterns,
  cooper2014multimodal}.  Likewise,  fMRI has also  proven informative
for    mental   illness    discrimination   \cite{gaebler2015auditory,
  turner2012reliability, liu2013abnormal}.   Therefore, we hypothesize
that given  each modality is  informative in a  complementary fashion,
the combination  of two  or more modalities  may increase  accuracy of
mental illness prediction.

While the richness of the  information embedded in brain data provides
great   promise   for   unveiling  hidden   features   and   knowledge
\cite{sui2012review}, it  also raises  a pressing  question on  how to
systemically   analyze    the   data   to   maximize    the   benefits
\cite{liu2014review,   meyer2012future}.   Some   of  the   challenges
inherent to multimodal data analysis are:
\begin{itemize}
\item   Different   data   modalities  are   often   incompatible   in
  dimensionality and  inherent physical  properties of the  data.  For
  example, sMRI is a three-dimensional  image while fMRI is a sequence
  of three-dimensional images.  This  makes the data incompatible with
  classical  data analysis  techniques such  as correlation  or linear
  regression.
\item Brain imaging data is  highly dimensional and scarce in samples.
  For instance, brain structure measured with sMRI can present 300,000
  or more intracranial voxels, while  the number of samples (subjects)
  collected     ranges      between     less     than      100     and
  2,800~\cite{sabuncu2014clinical}.
\item Most current data mining  methods for multimodal analysis assume
  linear  interaction between  modalities.   However, the  independent
  nature of each modality weakens  this assumption and compromises the
  reliability of the results.
\end{itemize}

In order to address the first problem, we first perform a segmentation
of  the sMRI  image to  produce a  gray matter  map and  summarize the
four-dimensional  fMRI  image  to   a  three-dimensional  image  using
amplitude    of    low     frequency    fluctuations    (ALFF)    maps
\cite{yu2007altered}.  This  is thus  a feature-based  fusion approach
\cite{calhoun2009feature}.  Even though  the interpretability  of ALFF
maps is not as direct as  sMRI, ALFF has proven informative and useful
for analysis of brain behavior \cite{turner2012reliability}.

We use synthetic data  generators \cite{ulloa2015synthetic} to address
the second problem.  The  generators feed machine learning classifiers
in an online  fashion.  This will allow the model  to learn from large
number of synthetic  samples, and enhance the  model predictive power.
We  then  propose  a  multimodal multi-layer  perceptron  (MLP)  model
pre-trained at  the input layer  with an independent MLP  on synthetic
data.  The MLP model also addresses the third challenge because of the
non-linear nature of MLPs.

To the best of our knowledge, we present the first study on multimodal
classification  with  a  MLP,  pre-trained with  synthetic  data,  and
applied  to   sMRI  and  fMRI  modalities   to  predict  schizophrenia
diagnosis.

\section{Materials and Methods}

In this  section, we present the  dataset used for our  experiments as
well as a detailed description  of the MLP architecture, including the
synthetic  data  generators,  used  for  prediction  of  schizophrenia
diagnosis using the sMRI and fMRI contained in the dataset.

\subsection{Dataset} \label{sec:dataset}

The dataset consists  of data collected from  multiple sites including
the University of California Irvine,  the University of California Los
Angeles, the University of  California San Francisco, Duke University,
University of North Carolina, University  of New Mexico, University of
Iowa, and University of Minnesota Institutional Review Boards.

\subsubsection{Participants}

Both  the  sMRI  and  fMRI  data  collected  belong  to  the  function
biomedical informatics  research network (fBIRN) dataset  as described
in \cite{segall2009voxel, turner2013multi}.  The number of subjects in
the fBIRN dataset,  after purging subjects without both  sMRI and fMRI
data, includes 135 schizophrenia patients (including schizophrenia and
schizoaffective disorder) and 169 healthy controls.

As described in \cite{turner2013multi}, the schizophrenia patients and
healthy  controls were  matched  as  much as  possible  for age,  sex,
handedness, and  race distributions, recruited from  eight sites, that
participated in  the study.   Each individual was  diagnosed following
the  Structured  Clinical  Interview  for DSM-IVTR  Axis  I  Disorders
(SCID-I/P)  \cite{first2002structured}. All  patients were  clinically
stable on anti-psychotic medication for at  least 2 months, and had an
illness duration of more than 1 year.

\subsubsection{MRI parameters}

The sMRI data was  set to a slice thickness of  1.2 mm, sagittal slice
orientation, and re-sliced to $2 \times 2 \times 2$ mm. The latter was
not a requirement  but set for convenience.  SPM5 was  used to segment
the brain into  white matter (WM), GM, and cerebral  spinal fluid with
unmodulated normalized parameters

The  imaging  protocol  for  the  fMRI   scans  at  all  sites  was  a
T2*-weighted  AC-PC  aligned  echo   planar  imaging  sequence  (TR/TE
2s/30ms, flip angle 77 degrees,  32 slices collected sequentially from
superior to inferior, $3.4 \times 3.4 \times  4$ mm with 1 mm gap, 162
frames, 5:38 min).  For the  resting scan, subjects were instructed to
lie still with eyes closed.

Further details on MRI settings  and pre-processing can be obtained in
\cite{segall2009voxel, turner2013multi}.

\subsubsection{Quality control}

For quality control,  each sMRI volume was correlated  with all others
to compute  the mean correlation as  a quality metric.  Images  with a
quality metric 2  standard deviations below the  mean were categorized
as noisy.  Nine images were discarded based on this analysis, yielding
a final dataset composed of 290 images.

\subsection{Multi-layer perceptron}
\label{sec:mlp}

A multilayer perceptron  (MLP) is a feed-forward  neural network model
that  projects   a  set  of   inputs  through  a  set   of  non-linear
operations. The model is trained  by reducing the binary cross-entropy
between    the   true    and   estimated    labels.   We    used   the
AdaGrad~\cite{zeiler2012adadelta} learning  algorithm to  optimize the
cost function. 

We designed three  MLPs, two for unimodal  pre-training with synthetic
data  generator and  one for  the  multimodal MLP  that combines  both
modalities by concatenation of hidden units.

The unimodal MLP is designed with 3 layers, where aside form the input
layer we set  20 hidden nodes at each other  layer, sigmoid activation
functions, 50\% dropout, and $L_2$ regularization with a weight of 0.1
at the input layer and 0.01 for the other layers.

The multimodal  MLP is  designed with  3 layers  for each  modality, a
merging layer that concatenates hidden  units on the fourth layer, and
2  merged layers  for final  output.  Before  concatenation all  input
layers have  20 hidden units, and  after concatenation it has  40, 20,
and 1 for the output. The $L_2$ regularization weight is set to 0.1 at
the input layers and 0.01 for the rest.

\subsection{Synthetic data generator}

The synthetic data generator is set to produce synthetic sMRI and ALFF
maps. It starts by fitting  unlabeled data using independent component
analysis (ICA), then it passes the ICA parameters to a random variable
(RV) generator that  imitates the parameters to  generate more samples
with the same statistical properties.  As in our case, labeled data is
given  to  the  RV  generator  to  capture  two  sets  of  statistical
parameters,  one  for healthy  controls  and  other for  schizophrenia
patients. The following  sections present a brief  description of ICA,
describe two RV generators and the final method.

\subsubsection{Independent Component Analysis}
\label{sec:ICA}

ICA  is a  matrix factorization  technique in  which an  observed data
matrix $X$ is factored as
\begin{equation}
  \label{eq:ica} X_{n \times m} = A_{n \times c} S_{c \times m},
\end{equation} where, $A$ is the mixing matrix, $S$ the source matrix,
$n$ the  number of samples, $m$  the number of variables,  and $c$ the
number of sources.  When the number of sources, $c$, is unknown, as in
most  real world  problems, it  can be  estimated using  the criterion
defined  in  \cite{li2007estimating}.   This matrix  factorization  is
possible given  that the $c$  sources in $S$ are  mutually independent
and non-Gaussian.

When factorizing structural MRI and  ALFF maps, data is organized into
a subject $\times$ voxel matrix. The mixing matrix $A$  then represents the
subjects' loading patterns,  i.e., how each source  is weighted across
subjects,  and  the rows  of  $S$  represent  the sources,  which  are
weighted patterns of voxels.

\subsubsection{RV generator: Rejection sampling}
\label{sec:rejection}

Rejection sampling is  a well-known technique for  RV generation which
samples    from    complex    probability    distribution    functions
(PDF). However, it is only defined for one-dimensional RVs and given a
PDF in close form.  We use  the classical rejection sampling method on
the marginal distributions  of the multivariate RVs  and without prior
knowledge of the RV PDF.

Let $R\{f_x,M\}$  be a  random variable  (RV) generator  function that
receives     as    input     a     probability    density     function
$f_x(x):\mathbb{R}\rightarrow\mathbb{R},$ and the number of samples to
generate, $M$. The generated samples are randomly drawn from the input
PDF, $f_x$.

First, the generator function samples two RVs, $u \sim U(0,1),$ and $v
\sim  U(b_0,b_1),$   where  $U$  denotes   the  PDF  of   the  uniform
distribution and  $(b_0,b_1)$ denote the minimum  and maximum observed
sample. The method then accepts $v$  as a sample from $f_x$ given that
$u<f_x(v)$. This  procedure is  repeated until  the desired  number of
samples is obtained.

\subsubsection{RV generator: Multivariate Normal}
\label{sec:mvn}

We use  the sample mean and  sample covariance matrix from  the matrix
data  as  input  of  this  RV generator.   Then,  we  use  a  spectral
decomposition  approach  for  generating  multivariate  random  normal
samples.  Contrary  to a  rejection sampling generator,  this approach
accounts for  correlation structure  among the  RVs, however  it loses
generality for marginal distributions.

\subsubsection{Generator}

The data generator is based on two assumptions:
\begin{itemize}
\item The estimation  of ICA sources from the observed  data is a good
approximation of the true sources.
\item  A  group   of  individuals  with  a   common  diagnosis  shares
statistical   properties  that   are   reflected   in  their   loading
coefficients ($A$).
\end{itemize}

The data generator fits unlabeled data  with ICA, from which it passes
the estimated mixing matrix to a  RV generator of choice.  Then, given
the labels, the generator captures observed statistical parameters for
each  label  group  and  generates mixing  parameters  with  the  same
observed statistical  properties. The  new mixing parameters  are then
used to reconstruct using the sources estimated from unlabeled data. A
more detailed description follows.

Given the listed assumptions are  met, our generator first factors the
observed  dataset  $X$   into  $A$  and  $S$  as   described  in  sec.
\ref{sec:ICA}.  Then, it splits  $A$ into sub-matrices $A_\mathrm{HC}$
and   $A_\mathrm{SZ}$,   which    represent   healthy   controls   and
schizophrenia patients respectively.  Next,  the method feeds each $A$
matrix  to  an  RV  generator  of choice,  as  described  in  sections
\ref{sec:rejection} or \ref{sec:mvn}.

In the  case of using the  rejection sampling method, we  estimate the
probability density functions (PDF) of each column of $A$ as follows
$$ f^i_x = \mathrm{pdf}_N \{ [A_\mathrm{HC}]_i \}, g^i_x = \mathrm{pdf}_N \{[A_\mathrm{SZ}]_i \},$$
where,  $i$ indicates  the  $i^{th}$  column of  the  matrix, and  the
$\mathrm{pdf}_N\{\cdot\}$   function   denotes  a   N-bin   normalized
histogram. The algorithm then proceeds to input $f^i_x$ and $g^i_x$ to
the rejection sampling RV generator as follows
\begin{align*}  [\hat{A}_\mathrm{HC}]_{M \times  c} &=  [R\{f^1_x,M\},
R\{f^2_x,M\}, ..., R\{f^c_x,M\} ]\\ [\hat{A}_\mathrm{SZ}]_{M \times c}
&= [R\{g^1_x,M\}, R\{g^2_x,M\}, ..., R\{g^c_x,M\} ],
\end{align*}  where,  $R\{\cdot\}$  is  the RV  generator  defined  in
\ref{sec:rejection}, and $\hat{A}$ denotes a synthetic mixing matrix.

In  the case  of the  multivariate normal  sampling method,  we simply
estimate  the  mean  and  covariance  matrix  of  $A_\mathrm{HC}$  and
$A_\mathrm{SZ},$  and   generate  $M$  samples  using   the  estimated
parameters,
$\hat{A}_{\mathrm{HC}}     \sim    \mathrm{MVN}(\bar{A}_{\mathrm{HC}},
\Sigma_{\mathrm{HC}}),$
and
$\hat{A}_{\mathrm{SZ}}     \sim    \mathrm{MVN}(\bar{A}_{\mathrm{SZ}},
\Sigma_{\mathrm{SZ}}).$

Finally, we reconstruct $M$ images for each diagnosis group by
\begin{align*}      [\hat{X}_\mathrm{HC}]_{M     \times      m}     &=
[\hat{A}_\mathrm{HC}]_{M  \times   c}  [S]_{c  \times   m}  +\bar{X}\\
[\hat{X}_\mathrm{SZ}]_{M \times m}  &= [\hat{A}_\mathrm{SZ}]_{M \times
c} [S]_{c \times m} +\bar{X},
\end{align*}  where,  $\bar{X}$ is  the  voxel  mean computed  at  the
beginning  of the  method, and  $\hat{X}$ is  the resulting  simulated
image.

\subsection{Experimental Setup}

\begin{figure} \centering
    \includegraphics[width=0.9\columnwidth]{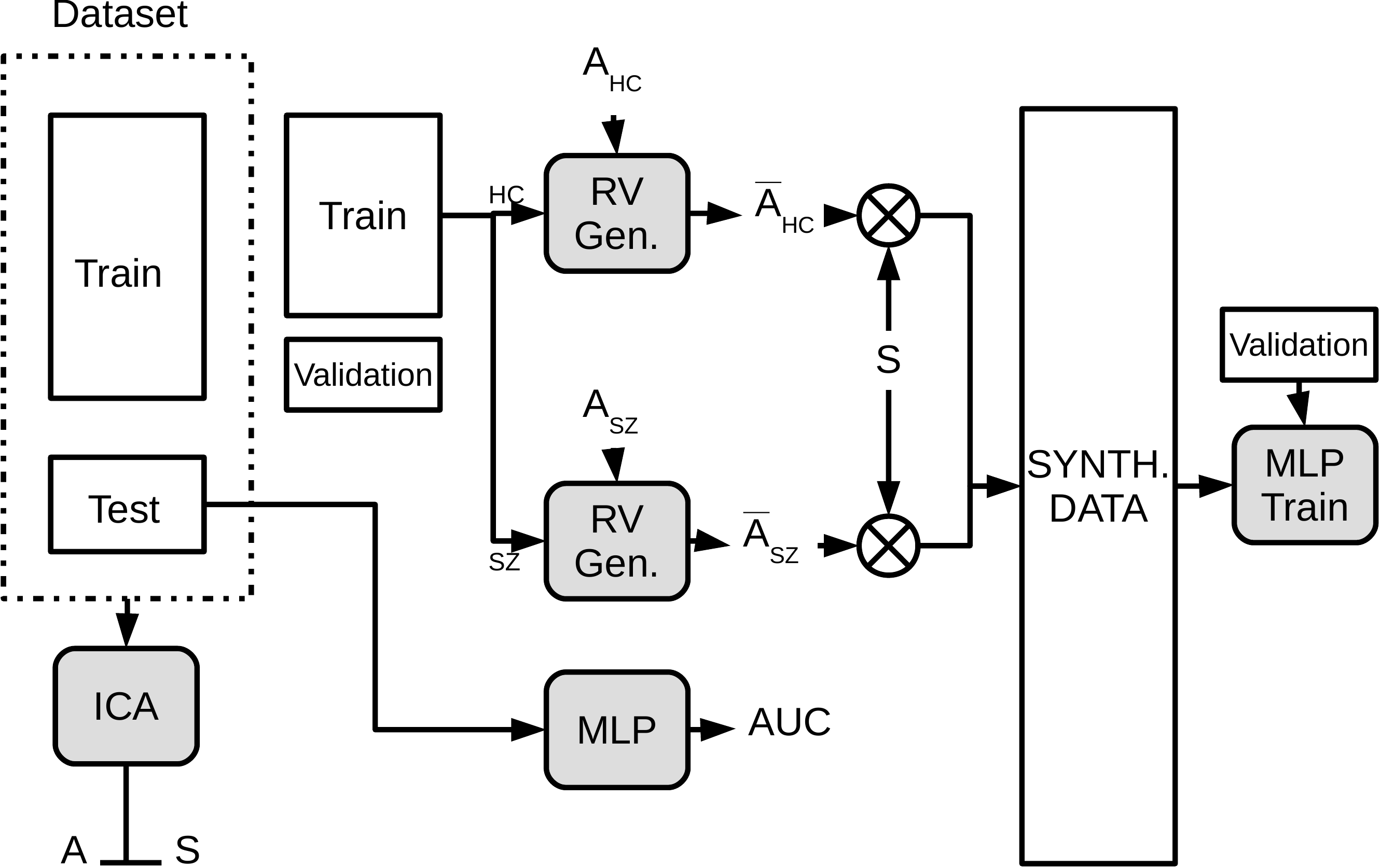}
    \caption{Experimental  setup for  unimodal MLP  training for  each
      fold.  The data is split into train and test datasets.  Then the
      training  set is  further  split into  training and  validation,
      which serves  as a  proxy of  testing data.   Each group  of the
      training data, healthy controls  (HC) and schizophrenia patients
      (SZ),  is passed  to the  RV generators  which form  a batch  of
      synthetic data to train a MLP.}
  \label{fig:uni-system}
\end{figure}

First, we fit the data generator for each modality using all unlabeled
data available.  This ensures we  estimate the best sources and mixing
matrix possible  in order  to meet  with the  first assumption  of the
generator.   Then, we  split the  data  of each  modality into  87.5\%
training and 12.5\% testing (8-fold cross-validation).

For  each data  modality,  the training  dataset is  fed  to the  data
generator, which  is set to produce  10,000 batches of 20  samples: 10
healthy  controls and  10 patients  with 20  estimated sources.   This
results  in  a  total  of  20,000  samples  per  modality  to  use  as
pre-training data. See  Fig.  \ref{fig:uni-system}  for a  view of  the
experimental setup for pre-training.

\begin{figure} \centering
    \includegraphics[width=0.9\columnwidth]{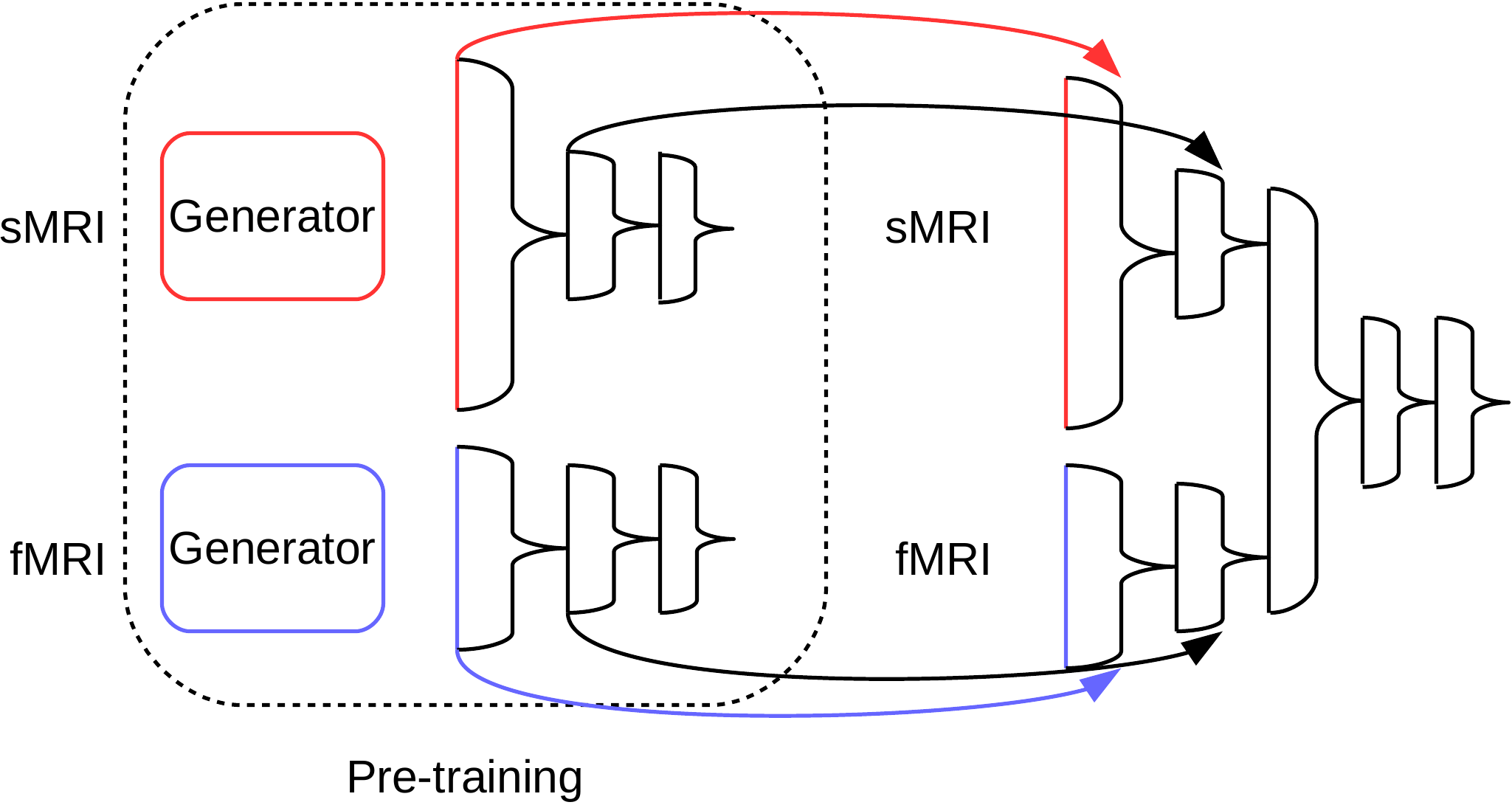}
    \caption{Multimodal MLP  model with  weight pre-training  for sMRI
      and  fMRI   data  modalities.  The  detailed   diagram  for  the
      pre-training is shown in Fig. \ref{fig:uni-system} }
  \label{fig:multi-system}
\end{figure}

Once each  unimodal MLP is pre-trained,  we use the input  weights and
set them  to initialize the  weight parameters of the  multimodal MLP.
The multimodal MLP  then starts training with real  training data. The
training procedure  is set  to split  90\% for  training and  10\% for
validation.  We use the validation dataset  as a proxy for the testing
set and avoid over fitting, thus, after 100 epochs we measure the loss
on validation  data and keep the  weights that results in  the minimum
validation loss after 1000 epochs. See Fig. \ref{fig:multi-system} for
a view of the complete system. 

In the pre-training phase, the method sequentially fed batches of data
to an online  trainer for a unimodal MLP classifier.   It is important
to notice that a sample of synthetic  data is only seen by the trained
model once, and  in practice the online learners are  fed with batches
sequentially without  first pre-generating the dataset  but generating
data  on-the-fly. 

We also train and test several other classical classifiers on raw data
for  comparison.   All  our  experiments are  implemented  using  free
software    provided     by    scikit-learn~\cite{scikit-learn}    and
Theano~\cite{theano2010}. Finally, we report  area under the ROC curve
(AUC) in the testing set.

For completeness, we run  classical classifiers including naive bayes,
logistic regression,  RBF and  Linear support vector  machines, linear
discriminant analysis, random forest,  nearest neighbors, and decision
tree. These  were run on raw  data for each modality  and concatenated
data  for  the  multimodal  approach.    We  used  a  grid  search  of
hyper-parameters and evaluated the best  combination within a nested a
10-fold  cross  validation,  we   then  report  average  and  standard
deviation of AUC across the 8 folds. For the concatenated approach, some
classifiers were too expensive to compute so we did not report results
as submission of this paper.

\section{Results}

The results indicate that the best classification results are obtained
when the information of sMRI and  fMRI is combined. Also, the proposed
MLP model  reported significantly higher  AUC average. See  a complete
summary of the results in Table \ref{tab:results}.

\begin{table}[]
  \caption{\label{tab:results} Average  and standard deviation  of the
area under  the ROC curve  (AUC) a 8-fold cross  validation experiment
for various classifiers and the proposed methodologies.}
\begin{center}
\begin{tabular}{@{}l|cc|cc|cc@{}} 
\toprule  & 
\multicolumn{2}{c}{ sMRI} & 
\multicolumn{2}{c}{ fMRI } & 
\multicolumn{2}{c}{sMRI + fMRI}\\
\begin{tabular}[c]{@{}l@{}}Classifier\\ Method\end{tabular} &
\begin{tabular}[c]{@{}c@{}}Average\\AUC\end{tabular} &
\begin{tabular}[c]{@{}c@{}}Standard\\ deviation\end{tabular} &
\begin{tabular}[c]{@{}c@{}}Average\\AUC\end{tabular} &
\begin{tabular}[c]{@{}c@{}}Standard\\ deviation\end{tabular} &
\begin{tabular}[c]{@{}c@{}}Average\\AUC\end{tabular} &
\begin{tabular}[c]{@{}c@{}}Standard\\     deviation\end{tabular}    \\
  \midrule  \multicolumn{7}{c}{Online learning  and  synthetic data}  \\
  \midrule 
  MLP with MVN         & 0.65 & 0.05 & 0.82 & 0.06 & \textbf{0.85} & \textbf{0.05} \\ 
  MLP with rejection   & \textbf{0.74} & \textbf{0.07} & \textbf{0.83} & \textbf{0.05} & 0.84 & 0.05 \\  \midrule
  \multicolumn{7}{c}{Raw data} \\\midrule 
  MLP                  & 0.65 & 0.09 & 0.82 & 0.10 & 0.80 & 0.08 \\
  Naive Bayes          & 0.62 & 0.10 & 0.71 & 0.11 & 0.61 & 0.07 \\ 
  Logistic Regression  & 0.69 & 0.12 & 0.82 & 0.07 & 0.81 & 0.08 \\ 
  RBF SVM              & 0.53 & 0.05 & 0.82 & 0.08 & 0.58 & 0.15 \\  
  Linear SVM           & 0.68 & 0.09 & 0.82 & 0.06 & 0.80 & 0.15 \\  
  LDA                  & 0.73 & 0.10 & 0.79 & 0.09 & 0.79 & 0.11 \\
  Random Forest        & 0.65 & 0.06 & 0.64 & 0.05 & 0.67 & 0.08 \\ 
  Nearest Neighbors    & 0.58 & 0.07 & 0.68 & 0.08 & 0.61 & 0.12 \\ 
  Decision Tree        & 0.56 & 0.11 & 0.54 & 0.10 & 0.53 & 0.13 \\\bottomrule
\end{tabular}
\end{center}
\end{table}

\section{Discussion}


We first  investigated the ability  of sMRI  and fMRI data  to predict
schizophrenia diagnosis. The results of various classifiers applied to
each individual  modality provides evidence that  both modalitites are
indeed   informative  since   the  overall   prediction  accuracy   is
significantly higher than random chance (0.5). The literature supports
our  findings in  both modalities  \cite{takayanagi2011classification,
  chyzhyk2015classification}, thus we can  assume with confidence that
sMRI and fMRI are of relevance for schizophrenia diagnosis.

Then,  we hypothesized  that given  sMRI and  fMRI hold  potential for
schizophrenia diagnosis prediction, the combination of both modalities
may improve  the overall classification accuracy.   Again, the results
show evidence in  favor of the stated hypothesis because,  as shown in
Table   \ref{tab:results},   the   proposed   multimodal   MLP   model
significantly  increased (p-value:  0.016, one-tailed  student t-test)
the average AUC  and reduced uncertainty among data  folds compared to
the best multimodal result with various classifiers.

Previous efforts on multimodal classification showed promising results
\cite{sui2012review}, yet, most of the  analysis is focused on feature
extraction under linearity  assumptions. In this study,  we proposed a
non-linear   approach,   MLP,    that   improves   generalization   in
classification  of schizophrenia  patients and  healthy controls  from
their sMRI and  fMRI images. Based on classification  results, the use
of the  proposed MLP model in  combination with the data  generator is
promising.

In general, multimodal deep learning  has gained popularity due to the
high    classification    rates    reported    in    the    literature
\cite{srivastava2012multimodal, kahou2015emonets}. However, all of the
application fields  are big data  problems.  This  is not the  case in
brain imaging.   The high cost  of MRI data collection  constrains the
amount of  data that can  be collected per  study.  In this  paper, we
used a synthetic data generation  technique to mitigate the effects of
a limited sample  size. As the results show, the  MLP model appears to
benefit from the  use of the generator which reduces  AUC variance and
slightly  improves classification  results compared  to the  MLP model
with out pre-training.

It is  well known that large  MLPs overfit the training  data, however
this seems to  not be the case for big  data problems, the overfitting
expected   from   nets   with    excess   capacity   did   not   occur
\cite{giles2001overfitting}.   Even though  our sample  size does  not
enter the  category of big data,  we are confident that  the synthetic
data generator used for pre-training played a role on regularizing the
unimodal  training.   The generator  was  used  to provide  more  than
200,000  samples   to  the   MLP  trainer.   Additionally,   to  avoid
overfitting  in   the  multimodal   training  phase,  we   used  other
regularization  methods, $L_2$  norm  and dropout,  besides using  the
pre-trained weights.

\section{Conclusion}

We  presented, to  our  knowledge,  the first  MLP  design using  data
generators  for  pre-training  applied  to  multimodal  brain  imaging
data. The use of the data  generator proved useful for pre-training in
the sense that it improved classification performance, probably acting
as a regularizer to avoid overfitting of the unimodal MLP model.

The multimodal design was a  simple concatenation of unimodal MLPs and
can be further used for more than two data modalities. As future work,
we could  assess the utility  of the multimodal MLP  including genetic
and behavioral information.


\bibliographystyle{IEEEtran} \bibliography{master}

\end{document}